\documentclass[journal]{IEEEtran}
\ifCLASSINFOpdf
\else
\fi
\usepackage{lineno,hyperref,amssymb}
\usepackage{graphicx}
\usepackage{algorithm}
\usepackage{algpseudocode}
\usepackage{threeparttable}
\usepackage{booktabs}
\usepackage{amsmath}
\usepackage{amssymb}

\begin{document}
%
\title{BackEISNN: A Deep Spiking Neural Network with Adaptive Self-Feedback and Balanced Excitatory-Inhibitory Neurons}
%
%
%

\author{Dongcheng Zhao, Yi Zeng, and Yang Li
\thanks{Dongcheng Zhao and Yang Li are with the Research Center for Brain-Inspired Intelligence,  Institute of Automation, Chinese Academy of Sciences, Beijing 100190, China, and also with the School of Artificial Intelligence, University of Chinese Academy of Sciences, Beijing 100049, China, (e-mail: zhaodongcheng2016@ia.ac.cn)}
\thanks{Yi Zeng is with the Research Center for Brain-Inspired Intelligence,  Institute of Automation, Chinese Academy of Sciences, Beijing 100190, China, also with the University of Chinese Academy of Sciences, Beijing 100049, and also with the Center for Excellence in Brain Science and Intelligence Technology, Chinese Academy of Sciences, Shanghai 200031, China (e-mail: yi.zeng@ia.ac.cn)} 
\thanks{The corresponding author is Yi Zeng.}
}

%
%

\markboth{Journal of \LaTeX\ Class Files,~Vol.~14, No.~8, February~2021}%
{Shell \MakeLowercase{\textit{et al.}}: Bare Demo of IEEEtran.cls for IEEE Journals}
%



\maketitle

\begin{abstract}
Spiking neural networks (SNNs) transmit information through discrete spikes, which performs well in processing spatial-temporal information. Due to the non-differentiable characteristic, there still exist difficulties in designing well-performed SNNs. Recently, SNNs trained with backpropagation have shown superior performance due to the proposal of the gradient approximation. However, the performance on complex tasks is still far away from the deep neural networks. Taking inspiration from the autapse in the brain which connects the spiking neurons with a self-feedback connection, we apply an adaptive time-delayed self-feedback on the membrane potential to regulate the spike precisions. As well as, we apply the balanced  excitatory and inhibitory neurons mechanism to control the spiking neurons' output dynamically. With the combination of the two mechanisms, we propose a deep spiking neural network with adaptive self-feedback and balanced excitatory and inhibitory neurons (BackEISNN). The experimental results on several standard datasets have shown that the two modules not only accelerate the convergence of the network but also improve the accuracy. For the MNIST, FashionMNIST, and N-MNIST datasets, our model has achieved state-of-the-art performance. For the CIFAR10 dataset, our BackEISNN also gets remarkable performance on a relatively light structure that competes against state-of-the-art SNNs.

\end{abstract}


%
\IEEEpeerreviewmaketitle

\section{Introduction}
In the past few years, deep neural networks (DNNs) have made tremendous progress in various machine learning tasks, such as computer vision~\cite{krizhevsky2017imagenet}, natural language processing~\cite{collobert2008unified}, and speech recognition~\cite{amodei2016deep}. However, DNNs only mimic the hierarchical topology structure of the information flow in the brain and process data in a real-valued form, far from the information-processing mechanisms in the brain. The spikes play a crucial role in efficient information processing. As a result, SNNs are developed to emulate the information processing in the brain. SNNs are considered the third-generation neural network~\cite{maass1997networks}, the discrete spike communication between the spiking neurons, multiple brain-inspired learning rules, and the intricate connections make it more biologically plausible and energy-efficient.

However, due to the lack of a effective training algorithm, the development of SNNs has stalled for a long time and has not yet demonstrated comparable performance to DNNs. Many researchers took inspiration from synapse learning in the brain, thus introduce some mechanisms that existed in the brains such as Spike-Timing-Dependent Plasticity (STDP)~\cite{bi1998synaptic}, Short-Term Facilitation (STF) or Depression (STD) into learning the weight of the SNNs~\cite{tavanaei2016bio,tavanaei2017multi,diehl2015unsupervised,falez2019multi,zhang2018plasticity,zhang2018brain}.    Although it has strong biological interpretability, most of these mechanisms are local learning rules, as the number of network layers goes more in-depth, it is difficult to achieve global convergence. Although GLSNN~\cite{zhao2020glsnn} introduced global feedback layers combined with local optimization learning rules, which alleviated this problem to a certain extent, it still performs poorly on complex datasets.

The success of deep learning comes from the proposal of the backpropagation (BP) algorithm. However, the discontinuous and non-differentiable charismatic of SNNs have impeded the BP implementation based on gradient descent. It's not feasible to apply the BP algorithm to the training of SNNs directly. An alternative method is to convert well-trained DNNs to SNNs through additional adjustments to parameters~\cite{diehl2015fast,xu2018csnn,sengupta2019going, Hu2018SpikingDR}. The conversion methods have achieved state-of-the-art accuracy on large datasets like ImageNet and complex architectures, such as VGG and ResNet. However, the superior performance comes from the well-trained DNNs, which does not solve the problem of SNNs' training. Also, the DNN-based training does not take full advantage of the temporal information in SNNs. When the SNNs go more in depth, the simulation length should be large enough to make the mean firing rates close to the real value of DNNs. Recently, due to the proposal of the approximation of the gradient of the spiking threshold functions, the backpropagation algorithm can be directly applied to the training of SNNs~\cite{lee2016training,wu2019direct,jin2018hybrid,wu2018spatio}. In this process, the discontinuous derivative of the spiking neurons is approximated by a continuous function.  Although they perform well on some simple datasets, there still exists a particular gap between them and the traditional DNNs.

In addition to the typical spiking neurons and the feedforward network structures, there are many other complicated mechanisms to support the brain's learning and inference. Here, we draw some mechanisms from the brain to further improve the BP-based SNNs. First, the typical network structure is based on simple forward structures. When the spiking neuron's membrane reaches the threshold, it releases a spike to the postsynaptic neuron. However, there are many other complex structures in the brain, such as the feedback connections~\cite{felleman1991distributed,sporns2004small}. The cross-layer feedback connections bring the information predicted by the higher cortex to the early cortical areas to help the inference and learning. Specially, the autapse  connected to the soma~\cite{ikeda2006autapses,wang2017formation,yin2018autapses} applies the time-delayed feedback on the neuron's membrane potential to regulate the spike precision and network activity.

As well, the excitatory spiking neurons are used in most of the current spiking neural networks. After the presynaptic neuron's membrane potential reaches the threshold, it releases the spike, increasing the membrane potential which makes the postsynaptic neurons easier fire. However,
there exist both excitatory and inhibitory neurons in the brain. The dynamic balance between excitatory and inhibitory neurons is crucial to healthy cognition and behavior~\cite{rubin2017balanced,dehghani2016dynamic}.

Taking inspiration from the two mechanisms, we introduce an adaptive time-delayed self-feedback mechanism to help regulate the membrane potential, which has improved the performance of the BP-based SNNs to some extent. Also, we introduce the inhibitory neurons in the BP-based SNNs. The dynamic balance of the excitatory and inhibitory neurons mechanism has accelerated the convergence of the neural networks and has improved the networks' performance. With the combination of the two mechanisms, we propose a deep spiking neural network with adaptive self-feedback and balanced excitatory inhibitory neurons (BackEISNN). And we have conducted experiments on several commonly used datasets to verify the performance of our BackEISNN. It has achieved state-of-the-art performance on MNIST, NMNIST, FashionMNIST datasets. Our contribution can be summarized as follows:
\begin{itemize}
\item We introduce the adaptive self-feedback mechanism (SFBM) to apply the time-delayed feedback on the membrane potential of the spiking neurons to regulate the spike precision. 
\item We introduce the balanced excitatory and inhibitory neurons mechanism (BEIM) to control the spikes firing in a balanced form, which has made the network achieve faster convergence and better performance.
\item Combined with the SFBM and BEIM, we have conducted extensive experiments on MNIST, Fashion-MNIST, NMNIST, CIFAR10 datasets. The results indicate that the proposed mechanisms could significantly improve the performance of the BP-based SNNs and accelerate the training of SNNs. We have achieved state-of-the-art performance on MNIST, NMNIST, and FashionMNIST datasets. And for the CIFAR10 dataset, we also get remarkable performance on a relatively light structure. 
\end{itemize}
\section{Related Work}
This section will review several BP-based SNNs in recent years and those which introduced some biological mechanism for better training.

The spikeprop~\cite{bohte2000spikeprop} algorithm could be viewed as the first attempt to train SNNs by operating on discontinuous spike activities. However, the single spike output restriction had limited it to demonstrate competitive performance on real-world tasks. DSN~\cite{o2016deep} proposed a spiking version of backpropagation to train a deep SNN with relu units. SCSNN~\cite{wu2019deep} treated the spike count as the surrogate for gradient backpropagation. BPSNN~\cite{lee2016training} treated the membrane potentials of spiking neurons as differentiable signals and treated spiking times as noise. 

However, all these methods lack explicit consideration of the temporal correlation of neural activities. To tackle the problems mentioned above, Slayer~\cite{shrestha2018slayer} handled the non-differentiable problem of the spike function by considering the temporal dependency between input and output signals. STBP~\cite{wu2018spatio} proposed spatial-temporal backpropagation for training SNNs, and the improved version~\cite{wu2019direct} introduced NeuNorm and decoding voting mechanisms, achieving higher performance on various datasets. HM2-BP~\cite{jin2018hybrid} proposed hybrid macro/micro level backpropagation. The micro-level was used to capture the temporal effects; both the macro and micro levels were used to compute the rate-coded errors. ST-RSBP~\cite{zhang2019spike} presented a spike-train level backpropagation algorithm to help train a recurrent spiking neural network.   LSNN~\cite{bellec2018long} introduced the adaptive neurons in the recurrent spiking neural networks to support the network acquire knowledge in a learning-to-learn scheme. TSSL-BP~\cite{zhang2020temporal} introduced two types of inter-neuron and intra-neuron dependencies of error backpropagation to improve the temporal learning precision.  The performance of the BP-based SNNs are still far behind that of DNNs due to the special spike transmission form. 

Others took inspiration from the learning of the brains from the structures and learning rules to further improve the learning.  LISNN~\cite{chenglisnn} introduced lateral interaction in SNNs, which enhanced the performance of the network and its robustness to noise to a certain extent. However, the performance was only verified on simple datasets. In addition to the STDP learning rules, Balance SNN~\cite{zhang2018brain} introduced the Short Term Plasticity (STP), and balance of excitatory-inhibitory neurons based on VPSNN~\cite{zhang2018plasticity}, but the proportion of excitability here was tested by manually trying different ratios. Also, winner-takes-all (WTA)~\cite{diehl2015unsupervised}, population coding~\cite{pan2019neural} were introduced into training SNNs. The performance gap still exists compared with the DNNs due to the special spike transmission form, which shows that we should take more inspiration from the brain's learning. Here, inspired by the autapse and the balanced excitatory and inhibitory neurons, we introduce the adaptive self-feedback mechanism and the balanced  excitatory and inhibitory neurons mechanism to help BP-based SNNs achieve better performance.

\section{Methods}
In this section, we will first give an introduction to the basic neuron model we used. And we will provide a detailed description of our SFBM and the BEIM. Finally, the whole pipeline of our BackEISNN will be presented.
\subsection{Basic LIF Neuron Model}
Leaky-Integrate-and-Fire (LIF) model is the most commonly used to describe the dynamic neural activities of the spiking neurons, including the dynamic changes of membrane potential and the firing process of spikes, which can be formulated as a differential Eq.~\ref{equa_lif1}:
\begin{equation}
 \tau\frac{dV(t)}{dt} = - V(t) + RI(t)
\label{equa_lif1}
\end{equation}

$V(t)$ is the membrane potential, $R$ is the membrane resistance, and $\tau = CR$ denotes the time constant. $I(t) = \sum_j^N w_{j, i}\delta_j$ denotes the total input generated by synaptic currents triggered by the arrival of spikes of presynaptic neurons. And when the membrane potential reaches the threshold $V_{th}$, the neuron will fire a spike.


For a better computational model, we modify Eq.~\ref{equa_lif1} in discrete form as shown in Eq.~\ref{equa_lif_2}:
 \begin{equation}
V_t = (1- \frac{1}{\tau})V_{t-1} + \frac{1}{C}I_t
\label{equa_lif_2}
\end{equation}

When the potentials fire a spike the membrane potential will be reset to $V_{reset}$. By assuming $V_{reset}=0$, $C=1$ as usual, the final equation can be written as:
 \begin{equation}
V_t = (1- \frac{1}{\tau})V_t(1 - \delta_{t-1}) + I_t
\label{equa_lif_3}
\end{equation}
$\delta_{t-1}$ is the spikes at time $t-1$.
When performing backpropagation in SNNs, the spike activation cannot be derived, which is a key issue that restricts the development of BP-trained SNNs. Here, we use the gradient of the rectangular function $S(V)$ to replace that of the original hard threshold function, which is also used in~\cite{wu2019direct,wu2018spatio}, the derivative of the rectangular function is shown in Eq.~\ref{equa_lif_4}:  
\begin{equation}
\frac{\partial S(V)}{\partial V}=\left\{
\begin{aligned}
1  & \qquad if \quad \frac{-1}{2} \leq V- V_{th}\geq \frac{1}{2} \\
0  & \qquad else \\
\end{aligned}
\right.
\label{equa_lif_4}
\end{equation}

\subsection{Adaptive Self-Feedback Mechanism}
\begin{figure}[htbp]
\centering
\includegraphics[width=8cm]{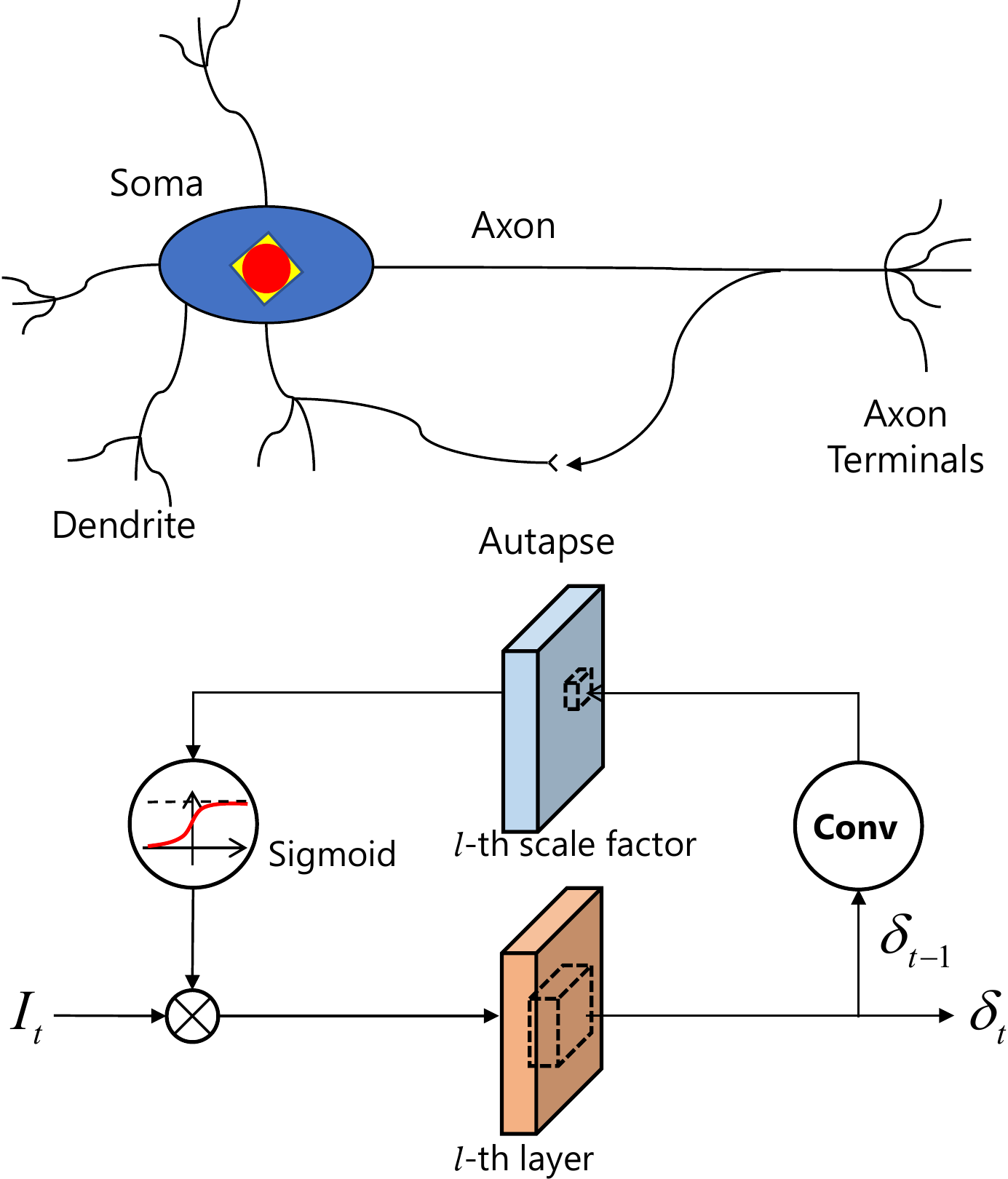}
\caption{The top is autapse in the brain, which connects the soma with a self-feedback manner. The bottom is the corresponding computational model.}
\label{Auta}
\end{figure}

In the brain, in addition to the traditional way of information transmission, presynaptic neuron emits spikes along the axons to the postsynaptic neurons. Specially, the autapse that connects the soma with a self-feedback loop that transmit information to itself. As shown in Fig.~\ref{Auta}, the autapses in the brain act on the membrane potential to help regulate the spike precision and network activity in a time-delayed self-feedback.

Inspired by the brain's autapses, we have introduced an adaptive self-feedback mechanism on the convolutional spiking neural networks. To reflect the modeling of the time-delay, we use the spikes fired at the last moment to regulate the input of the neurons. 

As shown in Fig.~\ref{Auta}, the spikes sent by the layer at the last moment were put into the convolution operation with a sigmoid activation function, which takes more advantage of the temporal information of SNNs. The output size is the same as the input current received by the layer, and the value is between 0 and 1. As a result, it can adjust the input adaptively to control the input to the membrane potential so that the spikes will fire more accurately. 


The detailed description of the adaptive self-feedback mechanism is shown in Eq.~\ref{loop1}, $f$ denotes the convolution operation, $\sigma$ denotes the sigmoid function, the self-feedback gate $SFB_t$ is multiplied to the input. $\odot$ means the element-wise.
\begin{equation}
\left
\{\begin{array}{l}
SFB_t = \sigma (f(\delta_{t-1}))\\
V_t = (1- \frac{1}{\tau})V_{t-1}(1 - \delta_{t-1}) + SFB_t \odot I_t\\
\end{array}\right.
\label{loop1}
\end{equation}
\begin{figure*}[htbp]
\centering
\includegraphics[width=18cm]{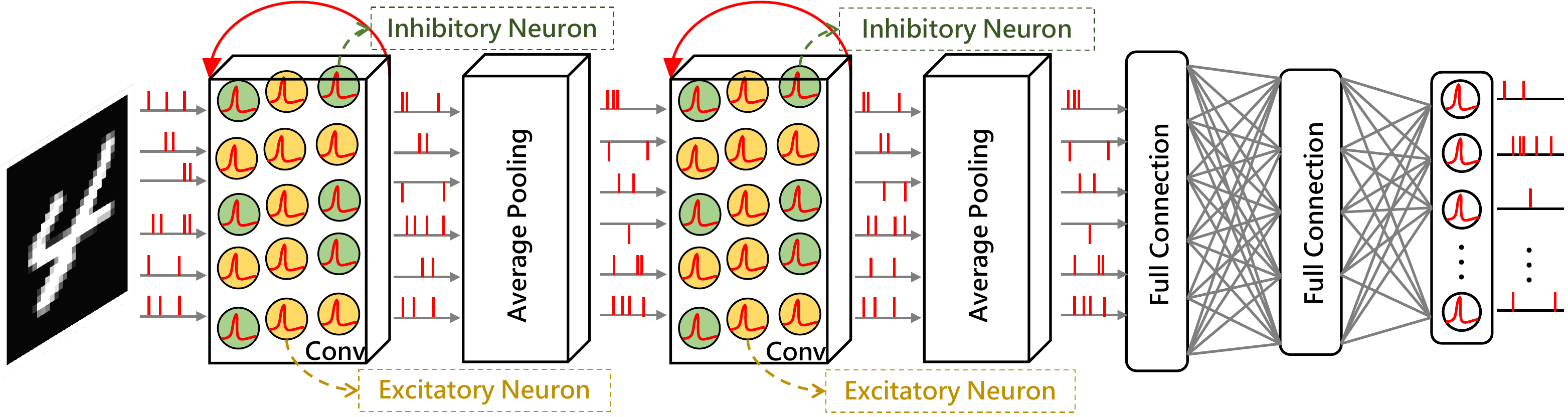}
\caption{The whole pipeline of our BackEISNN. The adaptive self-feedback mechanism is used to control the membrane to spike more precisely. The balanced excitatory and inhibitory mechanism is used to help the layer release spikes in a balanced manner.}
\label{lpnsnn}
\end{figure*}
\subsection{Balanced Excitatory and Inhibitory Neurons Mechanism}
\begin{figure}[htbp]
\centering
\includegraphics[width=8cm]{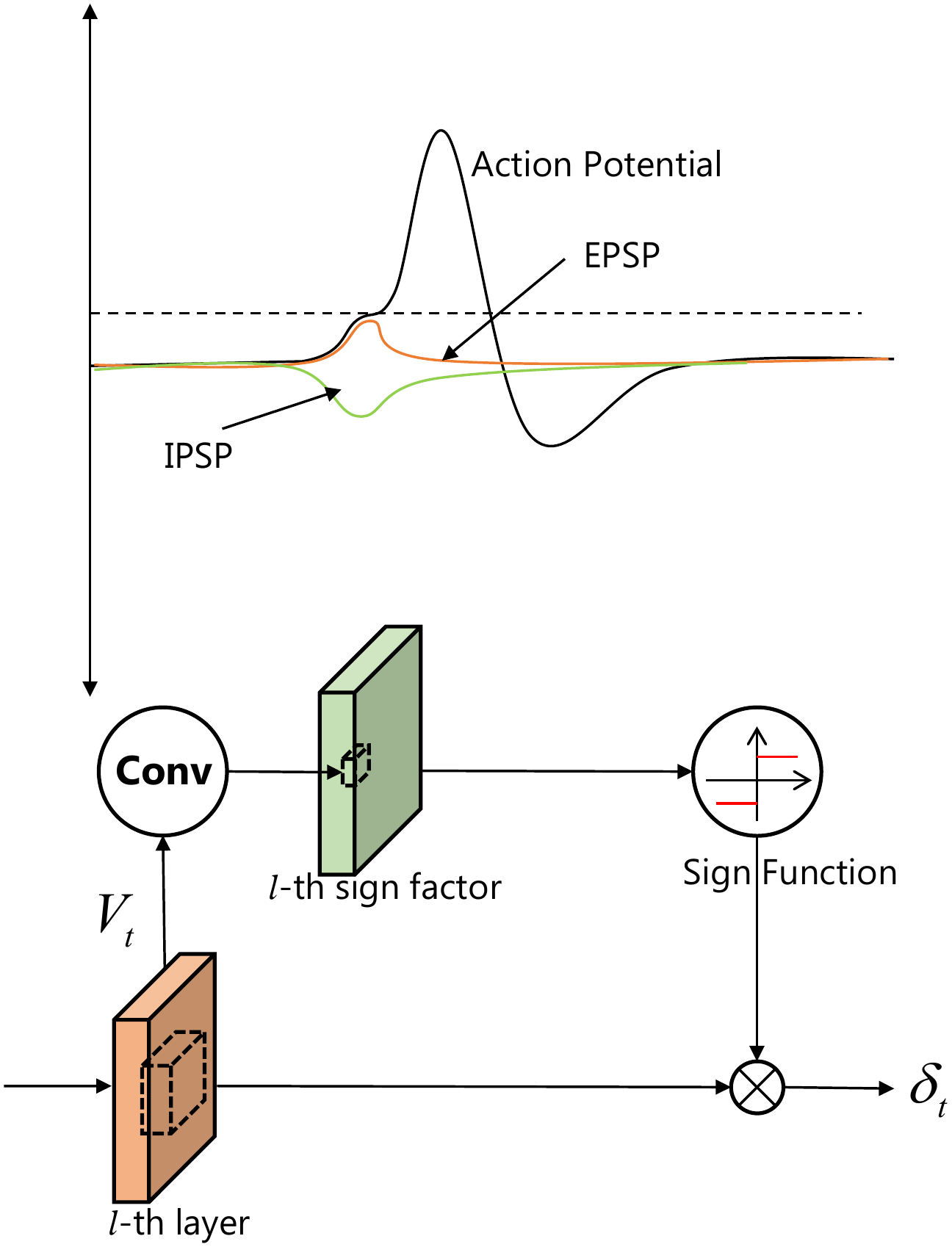}
\caption{The balance of the excitatory and inhibitory neurons in the brain and the corresponding computational model.}
\label{PN}
\end{figure}

There exist both excitatory and inhibitory neurons in the brain. The balanced combination of them helps the learning and inference. As can be seen in Fig.~\ref{PN}, the excitatory neurons exhibit the excitatory postsynaptic potential (EPSP), which allows the postsynaptic neuron easier to fire. In contrast, the inhibitory neurons exhibit the inhibitory postsynaptic potential (IPSP), making the postsynaptic neuron less likely to fire. We define excitatory neurons that release the positive spikes and inhibitory neurons that release negative spikes to simplify excitatory and inhibitory neurons' modeling. 

The positive combined with negative spikes were first used in SpikeGrad~\cite{thiele2019spikegrad}, which can be seen in Eq.~\ref{Ternary}:
\begin{equation}
S(V)=\left\{
\begin{aligned}
1  & \qquad if \quad V \geq V_{th} \\
-1  & \qquad if \quad V \leq -V_{th} \\
0  & \qquad else \\
\end{aligned}
\right.
\label{Ternary}
\end{equation}

The rules for sending positive or negative spikes are determined in advance. A positive spike occurs when the membrane potential exceeds the positive threshold, and a negative spike occurs when the membrane is below the negative threshold. While in the biological brain, the neuron would not fire a spike when the membrane potential does not reach the threshold. In this paper, we introduce a mechanism to determine whether the neurons are excitatory dynamically.  Inspired by Zhu’s work~\cite{zhu2011membrane} that the balance of excitation and inhibition is achieved at different membrane potentials of cortical neurons, we use the membrane potential to guide the generation of the balance.

As shown in Fig.~\ref{PN}, before the membrane potential is sent to the threshold function, it is sent into a convolutional layer with a sign activation function to determine whether the neurons are excitatory. And the gradient of this challenging function is processed the same as the hard threshold function mentioned above. When we get the neuron's positive and negative attributes, and the membrane potential reaches the threshold, we can decide whether to send a positive or negative spike.

The details can be seen in the Eq.~\ref{eqEI}, $f$ denotes the convolution operation, $\theta$ denotes the sign function, the excitatory and inhibitory neuron determining gate $EI_t$ is multiplied to the output.
\begin{equation}
\left
\{\begin{array}{l}
EI_t = \theta (f(V_t))\\
\delta_{t} = EI_t\odot S(V_t) 
\end{array}\right.
\label{eqEI}
\end{equation}

\subsection{The Whole BackEISNN Model}

The whole training process is shown in Fig.~\ref{lpnsnn}. The input was sent into the spiking convolutional networks and combined the two mechanisms mentioned above. The whole update process of the LIF neuron is shown in Eq.~\ref{all}:
\begin{equation}
\left
\{\begin{array}{l}
SFB_t = \sigma (f(\delta_{t-1}))\\
V_t = (1- \frac{1}{\tau})V_{t-1}(1 - \delta_{t-1}) + SFB_t \odot I_t\\
EI_t = \theta (f(V_t))\\
\delta_{t} = EI_t \odot S(V_t) 
\end{array}\right.
\label{all}
\end{equation}

Assuming that the output of the last layer is $O_t$ at time t, and for a given time window T, the mean average firing rates is $1/T\sum_{t=1}^T O_t$, $y_s$ is the true label, here we use the Mean Square Error loss function, as can be seen in Eq.~\ref{equa_lif_5}.
\begin{equation}
 L = \frac{1}{S} \sum_{s=1}^S ||y_s - 1/T\sum_{t=1}^T O_t|| _2^2
\label{equa_lif_5}
\end{equation}

\section{Experiments}
In this section, we perform the experiments on TITAN A100 GPU with PyTorch framework~\cite{paszke2017automatic}. We initialize the network weight with the default method of PyTorch. The optimizer is the Adam~\cite{kingma2014adam} optimizer. The batch size is set to 100. The epochs are set with 200. The learning rate $lr$ is set with 0.001 and decayed to $0.1lr$ every 40 epochs.  We set $V_{th}=0.5$. To illustrate the above two modules' improvement on the representation learning ability, we only apply the above two modules in the convolutional layers. We conducted the experiments on MNIST~\cite{lecun1998mnist}, N-MNIST~\cite{orchard2015converting}, Fashion MNIST~\cite{xiao2017fashion}, and CIFAR10~\cite{Krizhevsky09learningmultiple} datasets to illustrate the superiority of our proposed BackEISNN model.
\subsection{MNIST}
MNIST dataset is the most widely used classification dataset to measure the performance of deep learning models. It consists of 60,000 training examples and 10,000 test samples, describing the hand-written digits from 0 to 9. The shape of the sample is 28*28. For the static MNIST dataset, we first convert the samples into spike trains. The normalized pixel is converted to 1 or 0, comparing the pixel value with a random number between 0 to 1. If the value is greater than the random number, then a spike is triggered.
Our network structure is the same with ST-RSBP~\cite{zhang2019spike}, which used two convolutional layers and two linear layers. However, in our experiment, we did not use the data argumentation method elastic distortion. First, we fixed the simulation time to 20 and explored the impact of different kernel sizes of the two separate modules on the network results.

 \begin{figure}[htbp]
	\centering
	\includegraphics[width=7.5cm]{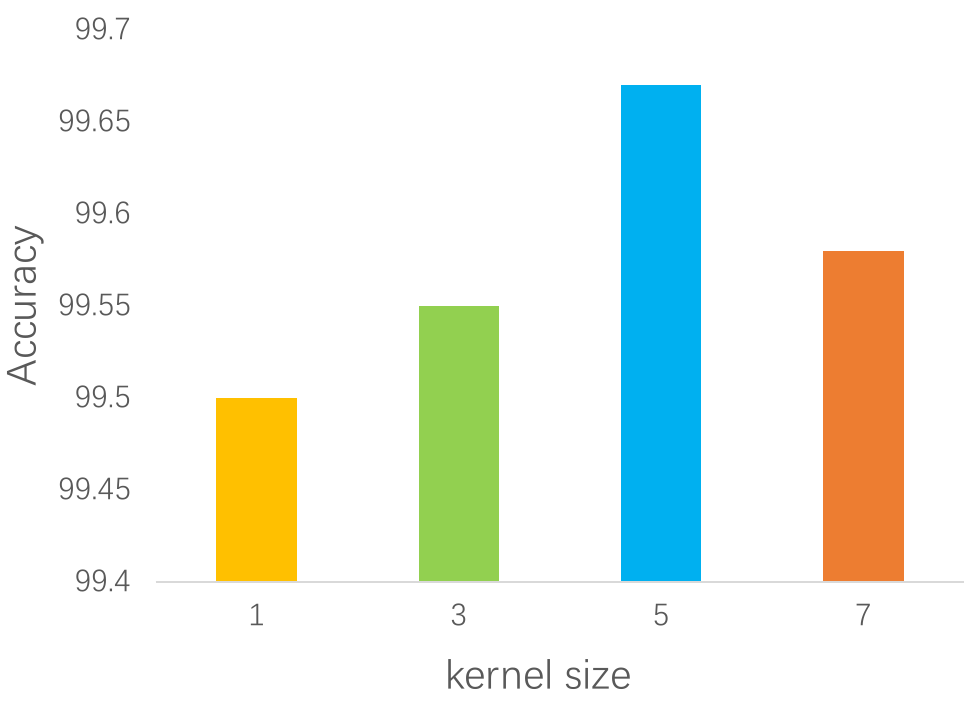}
	\caption{The test accuracy on MNIST with different kernel size of the two modules.}
	\label{kernel}
\end{figure}

As shown in Fig.~\ref{kernel}, when the kernel size is set with 5, the network achieves the best performance. Then we explore the impact of different simulation lengths on the network performance with the kernel size fixed at 5.
\begin{figure}[htbp]
	\centering
	\includegraphics[width=7.5cm]{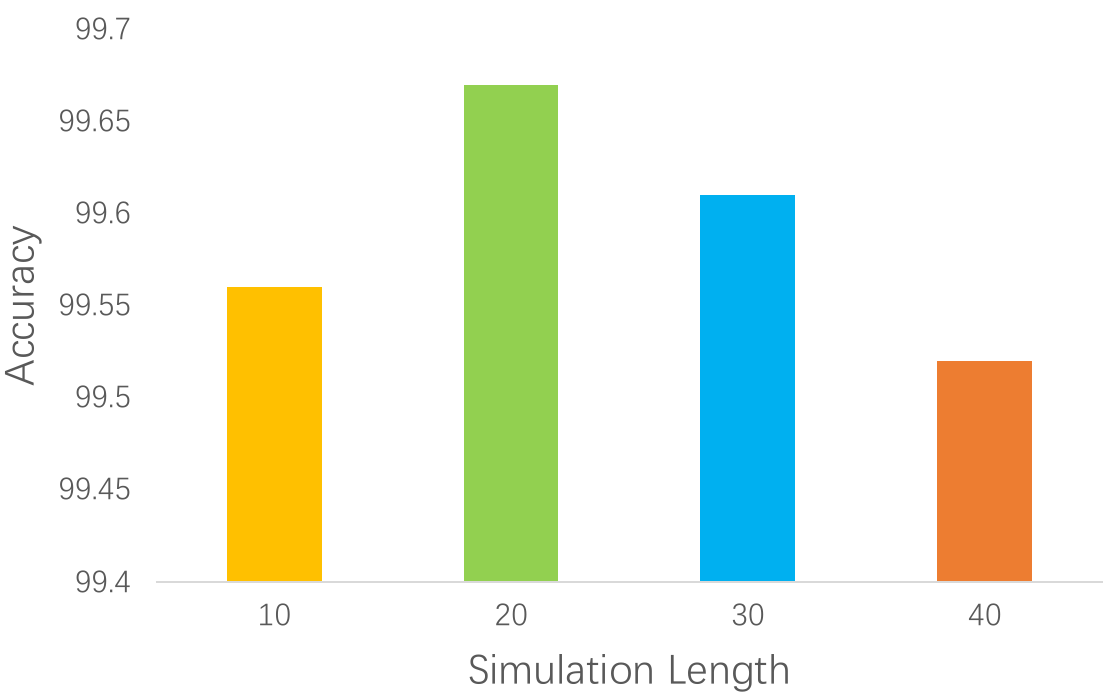}
	\caption{The test accuracy on MNIST with different simulation length.}
	\label{time}
\end{figure}

As shown in Fig.~\ref{time}, when the simulation time is set with 10, the network performance has reached 99.56\%, which has achieved relatively high performance when the simulation length is short. And when the simulation length is set with 20, the network reaches the best performance. 

\begin{figure}[htbp!]
	\centering
	\begin{minipage}{7.5cm}
		\includegraphics[width=7.5cm]{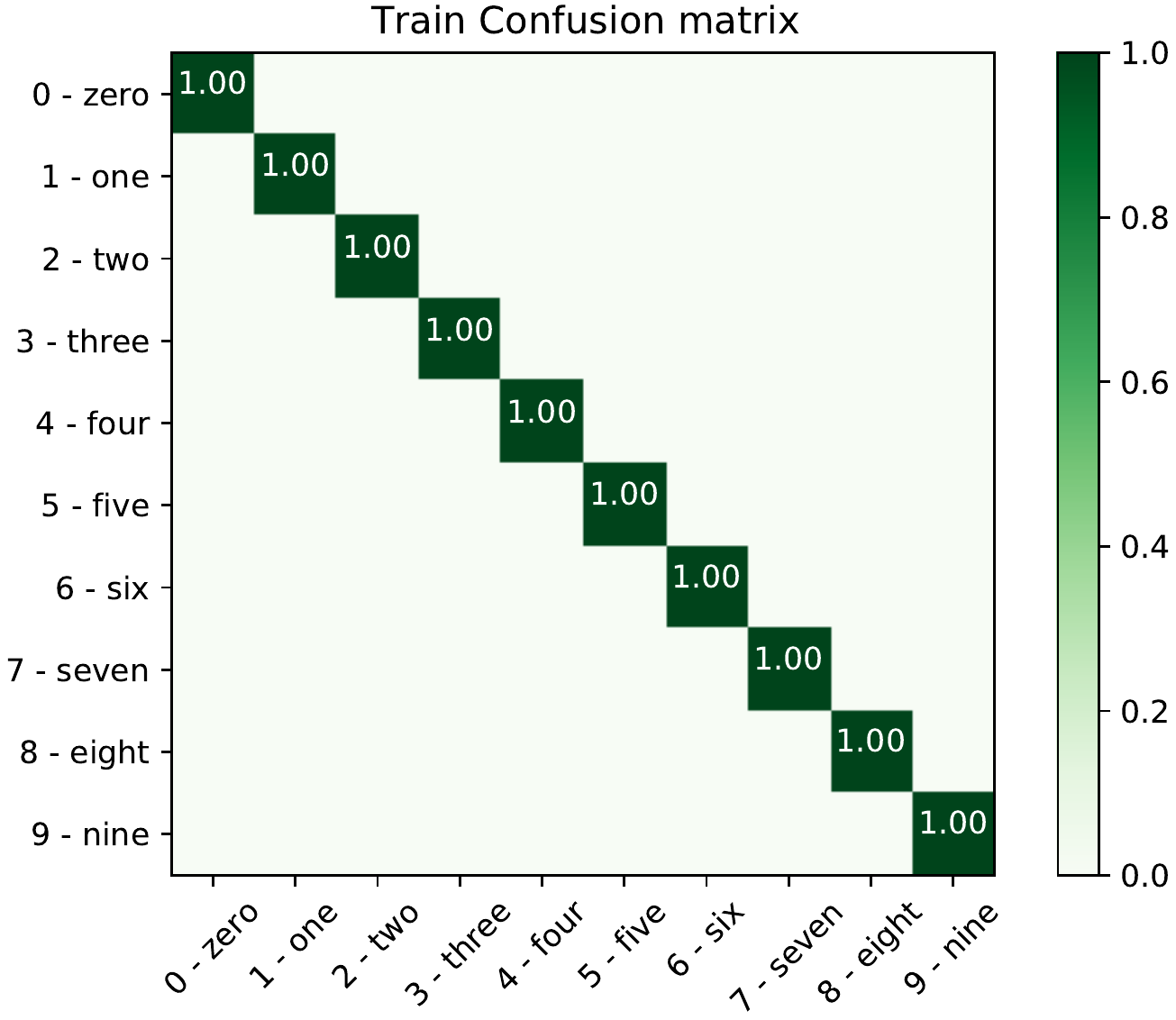}
	\end{minipage}
	\begin{minipage}{7.5cm}
		\includegraphics[width=7.5cm]{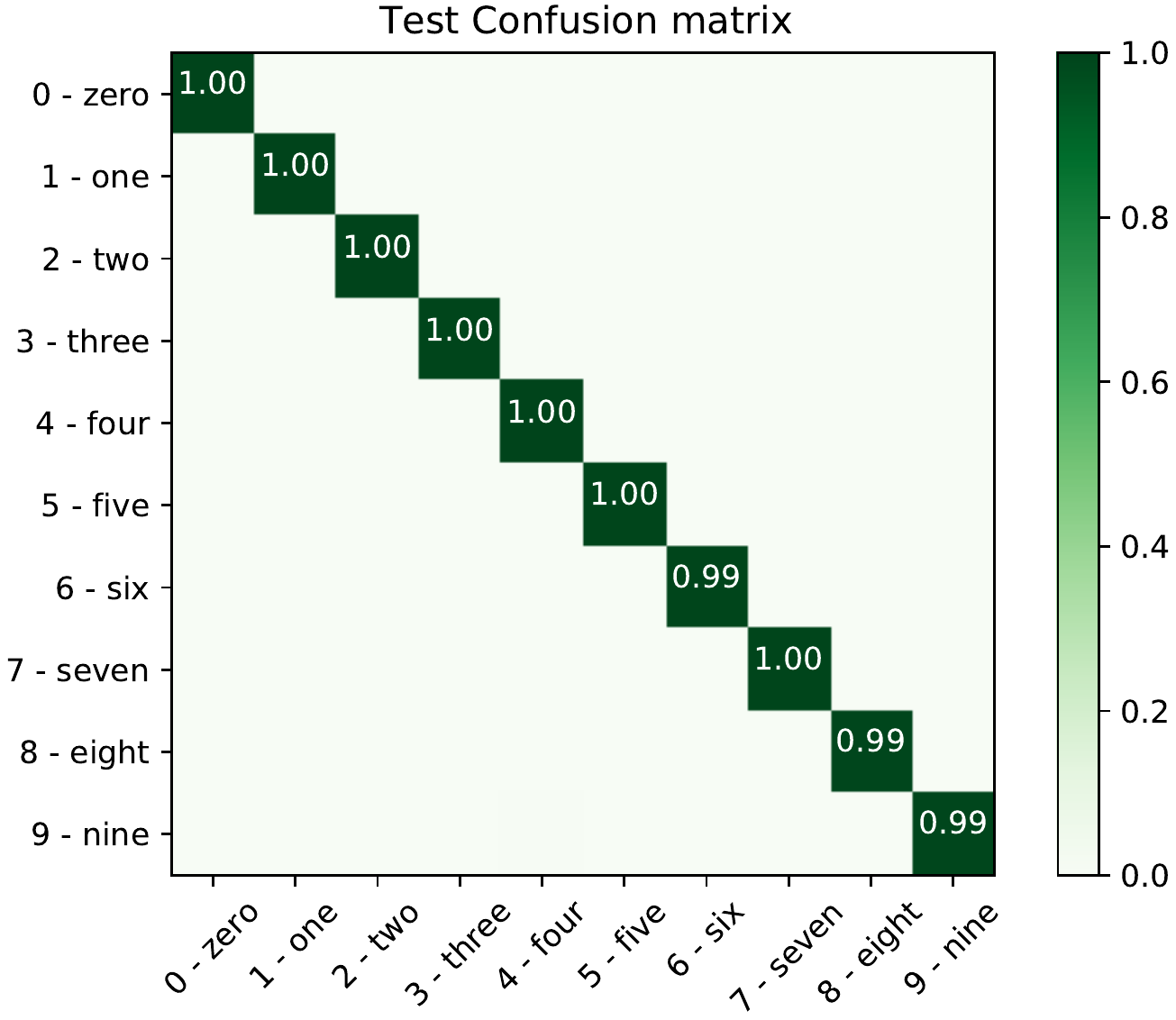}
	\end{minipage}
	\caption{Confusion matrix of the BackEISNN on the MNIST dataset. The top is the training dataset, and the bottom is the test dataset.}
	\label{confuse1}
\end{figure}

We have set the kernel size with 5 and the simulation length with 20. The experiment was repeated five times with different random number seeds. And the test accuracy is shown in Fig.~\ref{confuse1}. Each category is well classified. 

As shown in the Tab.~\ref{tab:addlabel}, several other BP-based SNNs were compared with our BackEISNN. Compared with STBP, which does not introduce the SFBM and the BEIM, we have an improvement of 0.25\%. For other BP-based SNN frameworks, our EILoopSNN has exceeded all of them, which is the state-of-the-art for the MNIST dataset.
 
\begin{table}[H]
	\centering
	\caption{Performances of BP based SNNs on MNIST}
	\begin{tabular}{cccccc}
		\toprule
		Model & Structure & Mean & Stddev &Best \\
		\midrule
		Spiking CNN~\cite{lee2016training} &20C5-P2-50C5-P2-200 &- & -
   & 99.31 \\
   SLAYER~\cite{shrestha2018slayer} &12C5-P2-64C5-p2 &99.36 & 0.05
   & 99.41 \\
		STBP~\cite{wu2018spatio} &15C5-P2-40C5-p2-300 & - & -
    & 99.42 \\
		HM2BP~\cite{jin2018hybrid}
 &15C5-P2-40C5-p2-300 & 99.42 & 0.11 
    & 99.49 \\
    LISNN~\cite{chenglisnn}
 &32C3-P2-32C3-P2-128 & - & -
    & 99.5 \\
     TSSL-BP~\cite{zhang2020temporal} 
 &15C5-P2-40C5-p2-300& 99.5 & 0.02
    & 99.53 \\
		ST-RSBP~\cite{zhang2019spike}
  &15C5-P2-40C5-p2-300
 & 99.57
& 0.04
 & 99.62
 \\
     \textbf{BackEISNN (ours)}
 &15C5-P2-40C5-p2-300 & \textbf{99.58} & 0.06
    & \textbf{99.67} \\
		\bottomrule
	\end{tabular}%
	\label{tab:addlabel}%
\end{table}%

\subsection{N-MNIST}
The N-MNIST dataset is a neuromorphic version of the MNIST dataset by mounting a Dynamic Version Sensor (DVS) in front of static digit images on a computer screen. It has 60,000 training samples and 10,000 test samples, which is the same as MNIST. The DVS is moved in the direction of three sides of the isosceles triangle in turn and collects the two-channel spike event (since the intensity can be increased or decreased) triggered by the pixel change lasting for 300ms. Due to each image's relative shifts, each sample of the N-MNSIT is a spatial-temporal pattern with 34x34x2 spike sequences. Since the time resolution is 1us, there is a total of 300,000-time steps. To speed up the simulation, we reduced it to 100-time steps, which means no matter how many events during the 3000-time intervals, we only record one spike. We also test the influence of different kernel sizes, and as shown in Tab.~\ref{tab:kernelnmnist}, when the kernel size is set with 3, the network reaches the highest performance. 
\begin{table}[htbp]
\centering
\caption{The performance of network on N-MNSIT dataset with different kernel sizes}
\begin{tabular}{|c|c|c|c|c|c|}
\hline  
kernel size & kernel 1 & kernel 3 & kernel 5  & kernel 7 \\
\hline  
performance &99.44 & 99.57 & 98.89  & 96.55 \\
\hline  
\end{tabular}
\label{tab:kernelnmnist}%
\end{table}

\begin{figure}[htbp!]
	\centering
	\begin{minipage}{7.5cm}
		\includegraphics[width=7.5cm]{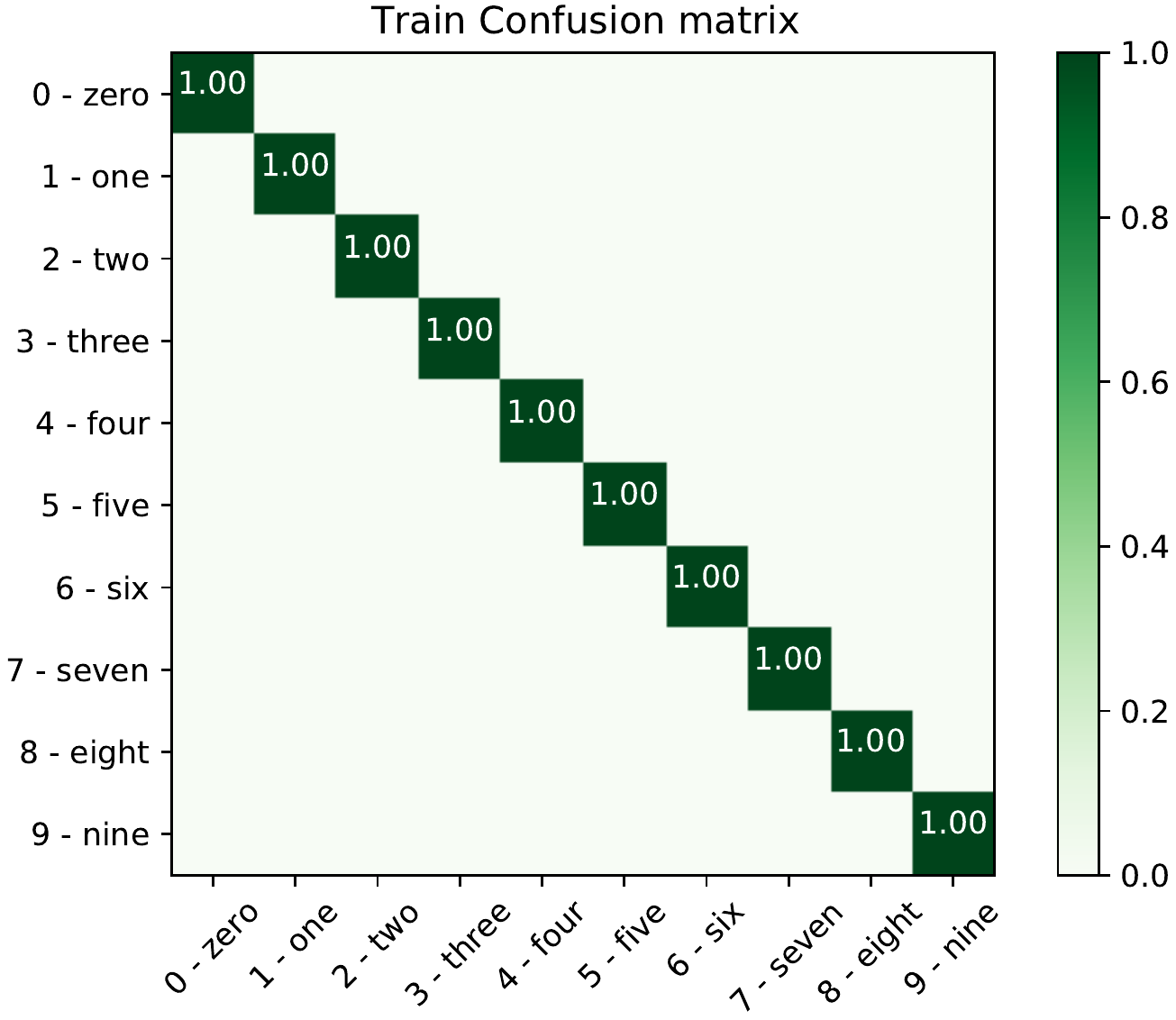}
	\end{minipage}
	\begin{minipage}{7.5cm}
		\includegraphics[width=7.5cm]{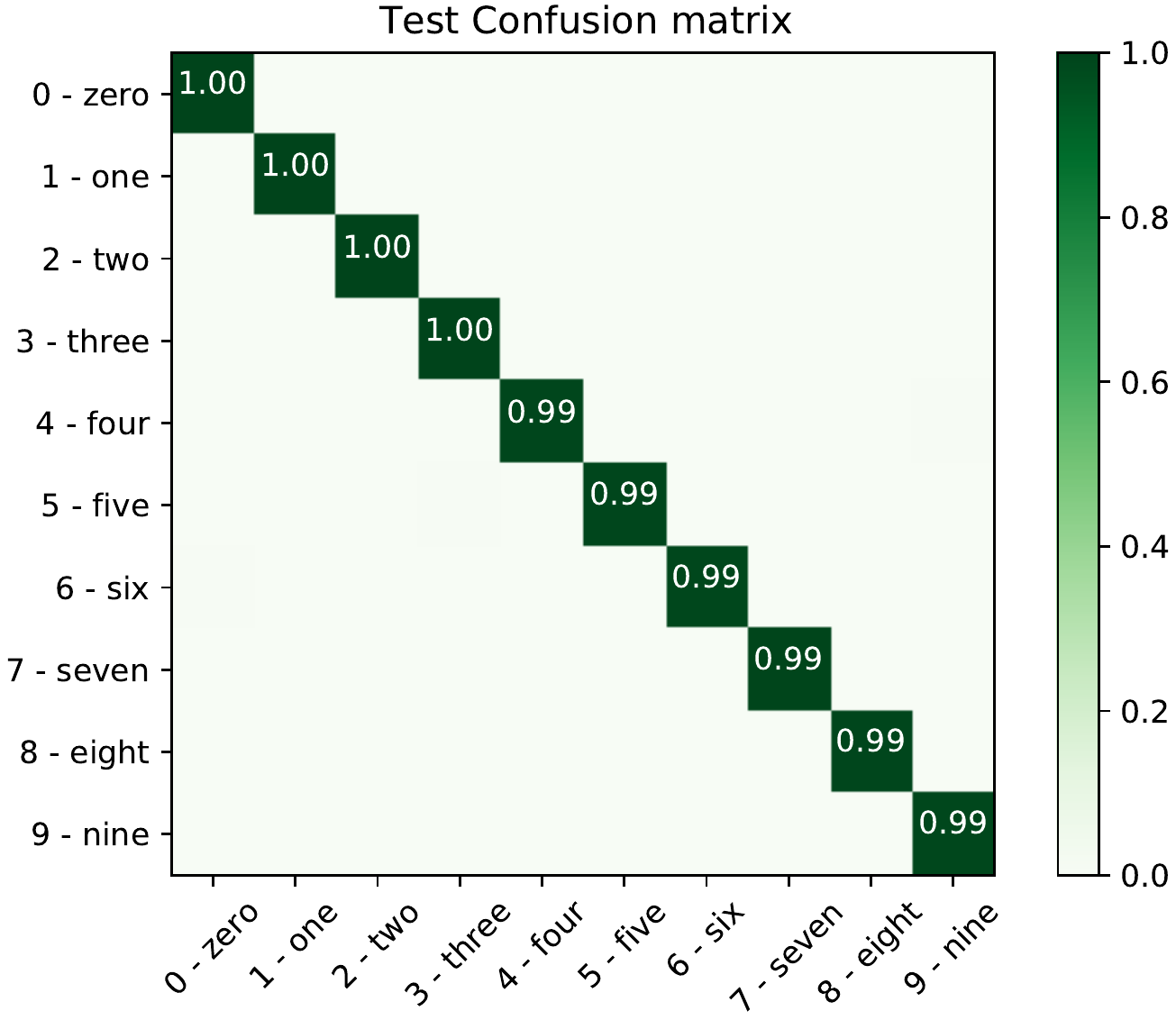}
	\end{minipage}
	\caption{Confusion matrix of the BackEISNN on the N-MNIST dataset. The top is the training dataset, and the bottom is the test dataset.}
	\label{ACC}
\end{figure}
The experiment is first conducted on 100-time steps. As can be seen in Tab.~\ref{tab:addlabe3}, our BackEISNN-100 has achieved 99.57\% accuracy and has achieved state-of-the-art performance compared with other BP-based SNNs. Moreover, we use the first 30 time-steps to experiment, and the result is denoted as BackEISNN-30, which still keeps the high level of performance.
\begin{table}[H]
	\centering
	\caption{Performances of BP based SNNs on N-MNIST}
	\begin{threeparttable}
	\begin{tabular}{cccccc}
		\toprule
		Model & Structure  & Mean & Stddev & Best\\
		\midrule
		HM2-BP~\cite{jin2018hybrid} &400-400 & - & - & 98.88 \\
		SLAYER~\cite{shrestha2018slayer} &Net 1 & - & - & 99.2 \\
		TSSL-BP 30~\cite{zhang2020temporal} &Net 1 & 99.23 & 0.05 & 99.28 \\
		TSSL-BP 100~\cite{zhang2020temporal} &Net 1 & 99.35 & 0.03 & 99.4 \\
		STBP~\cite{wu2019direct} & Net 2 & - & - & 99.44 \\
		LISNN~\cite{chenglisnn} &Net 3 & - & - & 99.45 \\
		STBP NeuNorm~\cite{wu2019direct} &Net 2  & - & - & 99.53 \\
   \textbf{BackEISNN-30 (Ours)} &Net 1& \textbf{99.47} & 0.05 & \textbf{99.56}\\
   \textbf{BackEISNN-100 (Ours)} &Net 1& \textbf{99.5} & 0.06 & \textbf{99.57}\\
		\bottomrule
	\end{tabular}%
	\label{tab:addlabe3}%
	        \begin{tablenotes}    
        \footnotesize               
        \item[1] Net 1 is 12C5-P2-64C5-P2
        \item[2] Net 2 is 128C3-128C3-P2-128C3-256C3-P2-1024-Voting
        \item[3] Net 3 is 32C3-P2-32C3-P2-128 
      \end{tablenotes}  
\end{threeparttable} 
\end{table}%

\subsection{FashionMNIST}
Fashion-MNIST is a complicated version compared to MNIST. It consists of gray-scale images of clothing experiments. As the same as the MNIST dataset process, the dataset is processed to spike trains.

First, we also analyze the impact of different kernel sizes on performance. As shown in the Tab~\ref{tab:kernel}, when the kernel size is set with 5, the network gets the best performance.
\begin{table}[htbp]
\centering
\caption{The performance of network on FashionMNSIT dataset with different kernel sizes in encoding manner.}
\begin{tabular}{|c|c|c|c|c|c|}
\hline  
kernel size & kernel 1 & kernel 3 & kernel 5  & kernel 7 \\
\hline  
performance &92.32 & 92.49 & 92.51  & 92.36 \\
\hline  
\end{tabular}
\label{tab:kernel}%
\end{table}

Then, with the kernel size fixed at 5, we analyze the influence of the simulation length. As can be seen in the Tab~\ref{tab:kernel2}, when the simulation length is 30, our BackEISNN reaches the best performance.
\begin{table}[htbp]
\centering
\caption{The performance of network on FashionMNSIT dataset with different simulation length in encoding manner.}
\begin{tabular}{|c|c|c|c|c|c|}
\hline  
simulation length & 10 & 20 & 30  & 40 \\
\hline  
performance &92.13 & 92.51 & 92.7  & 92.2 \\
\hline 
\end{tabular}
\label{tab:kernel2}%
\end{table}

For the TSSL-BP model, the real-value was directly inputted into the network. To be consistent with this approach, we also use the real-valued input current as of the direct input of the network to demonstrate the influence of the kernel size and the simulation length. We repeat the experiment as mentioned above. As shown in Tab.~\ref{tab:kernel3} and  Tab.~\ref{tab:kernel4} when the kernel size is 3 and the simulation length is 20, the network reaches the highest performance. 
\begin{table}[htbp]
\centering
\caption{The performance of network on FashionMNSIT dataset with different kernel sizes  in direct input manner.}
\begin{tabular}{|c|c|c|c|c|c|}
\hline  
kernel size & kernel 1 & kernel 3 & kernel 5  & kernel 7 \\
\hline  
performance &92.77 & 93.45 & 93.12  & 92.37 \\
\hline
\end{tabular}
\label{tab:kernel3}%
\end{table}

\begin{table}[htbp]
\centering
\caption{The performance of network on FashionMNSIT dataset with different simulation length in direct input manner.}
\begin{tabular}{|c|c|c|c|c|c|}
\hline
simulation length & 10 & 20 & 30  & 40 \\
\hline  
performance &92.86 & 93.45 & 93.35 & 93.11 \\
\hline  
\end{tabular}
\label{tab:kernel4}%
\end{table}

We repeat the experiments five times with the parameters mentioned above. The result of the encoding input is denoted as BackEISNN-E, while the direct input is denoted as BackEISNN-D. As shown in Tab.~\ref{tab:addlabe2}, for the encoding results, our BackEISNN-E has achieved 92.7\% accuracy and surpasses most of the results. For the TSSL-BP, it uses the direct input encoding, as shown in Tab.~\ref{tab:addlabe2}, our BackEISNN-D has achieved state-of-the-art performance on the FashionMNIST dataset, indicating our algorithm's progressive nature.

\begin{table}[H]
	\centering
	\caption{Performances of BP based SNNs on FashionMNIST}
	\begin{threeparttable}
	\begin{tabular}{cccccc}
		\toprule
		Model & Structure  & Mean & Stddev & Best\\
		\midrule
		HM2BP~\cite{jin2018hybrid} &400-400 & - & - & 88.99 \\
		GLSNN~\cite{zhao2020glsnn} &256*8& - & - & 89.02\\
		ST-RSBP~\cite{zhang2019spike} &400-R400& - & - & 90.13\\
		LISNN~\cite{chenglisnn} &Net 1& - & - & 92.07\\
		BackEISNN-E (ours) &Net 2& 92.56 & 0.08 & 92.7\\
   TSSL-BP~\cite{zhang2020temporal}  &Net 2& 92.69 & 0.09 & 92.83\\
   \textbf{BackEISNN-D (Ours)} &Net 2& \textbf{93.04} & 0.31 & \textbf{93.45}\\
		\bottomrule
	\end{tabular}%
	\label{tab:addlabe2}%
	\begin{tablenotes}    
        \footnotesize               
        \item[1] Net 1 is 32C3-P2-32C3-P2-128
        \item[2] Net 2 is 32C5-P2-64C5-p2-1024
      \end{tablenotes}  
\end{threeparttable} 
\end{table}%

\begin{figure}[htbp!]
	\centering
	\begin{minipage}{7.5cm}
		\includegraphics[width=7.5cm]{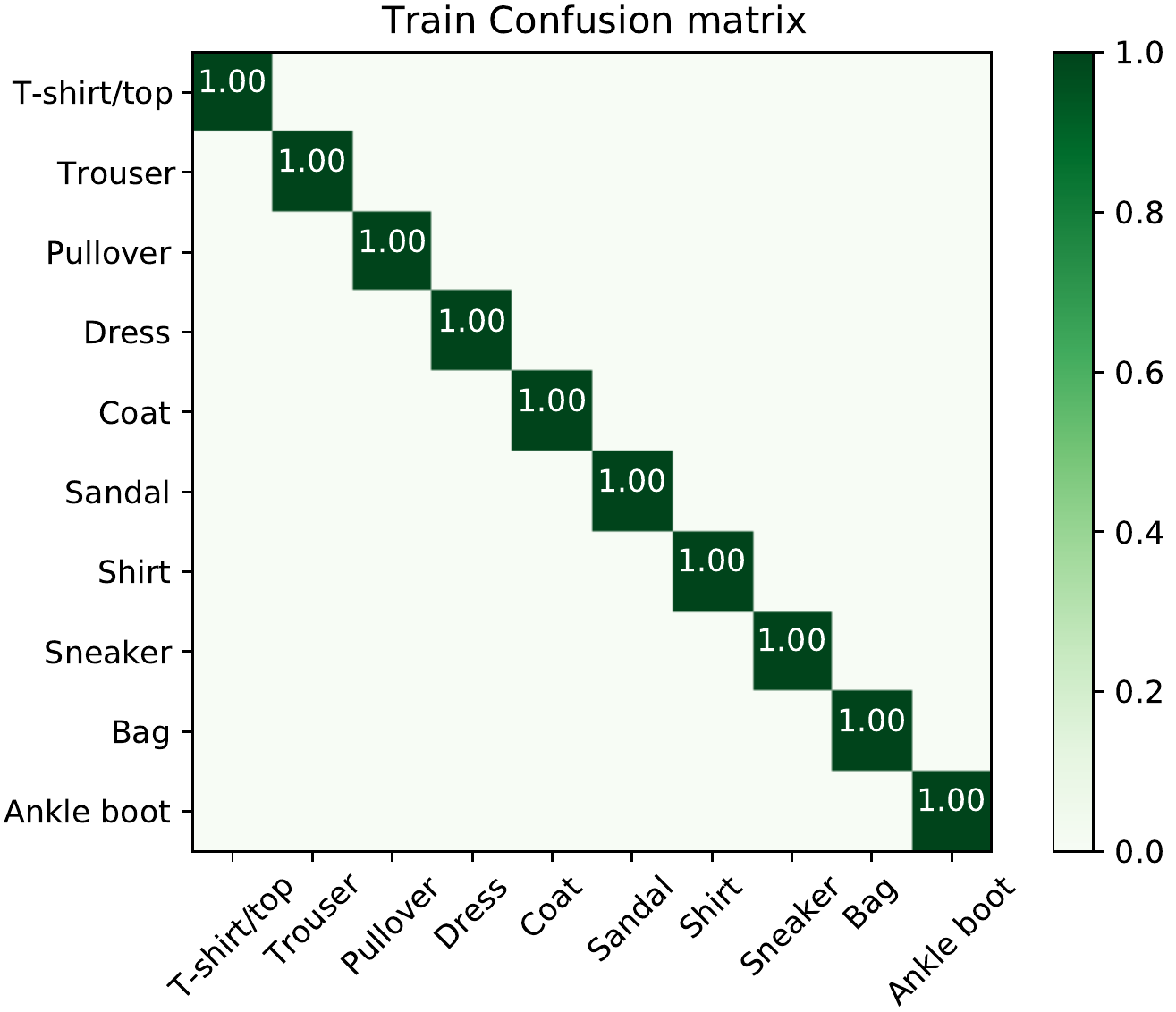}
	\end{minipage}
	\begin{minipage}{7.5cm}
		\includegraphics[width=7.5cm]{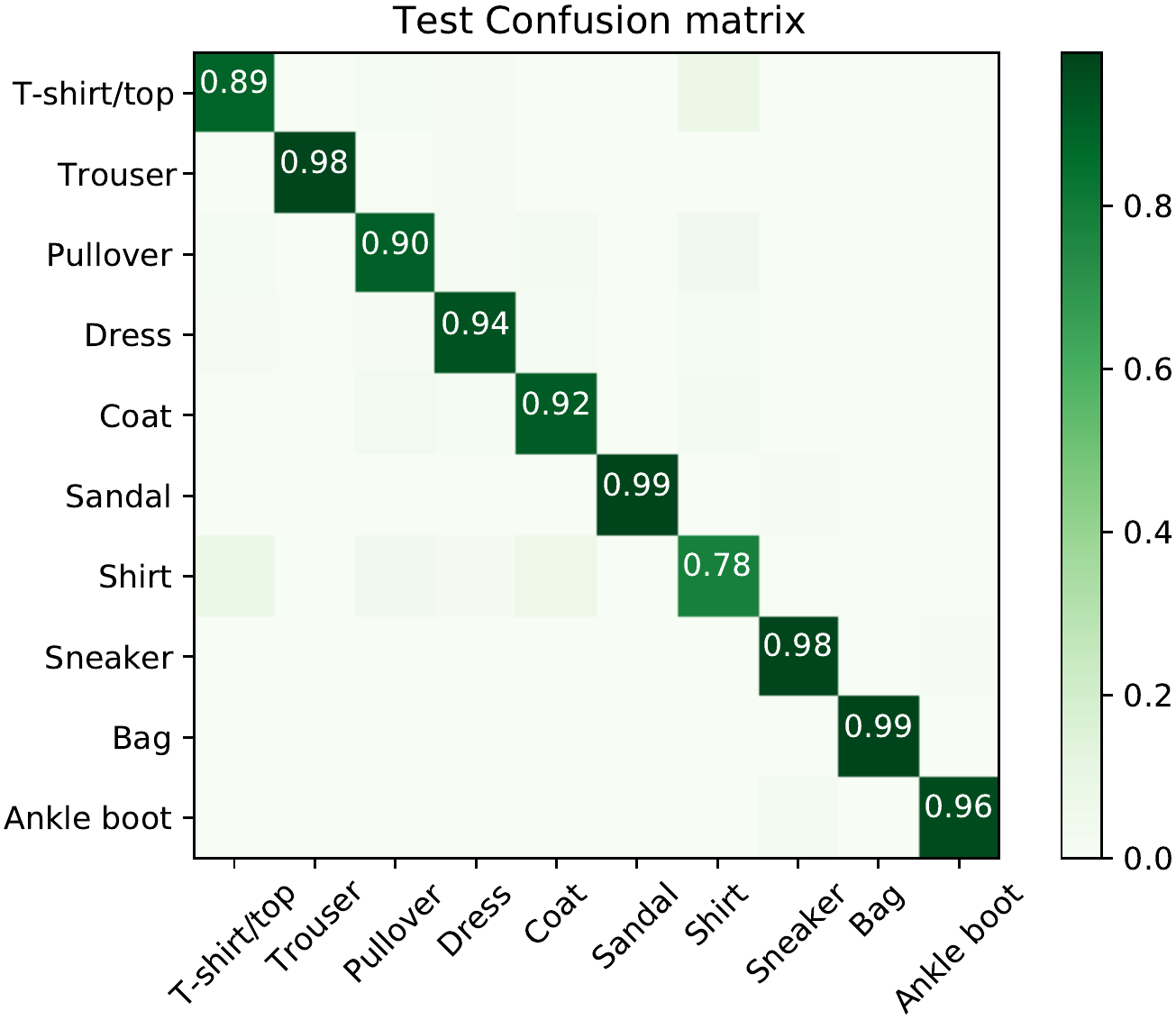}
	\end{minipage}
	\caption{Confusion matrix of the BackEISNN on the FashionMNIST dataset. The top is the training dataset, and the bottom is the test dataset.}
	\label{ACC}
\end{figure}
To better analyze the result, we plot the confusion matrix. As can be seen in Fig~\ref{ACC}, for the training dataset, our BackEISNN can distinguish each category clearly. However, for the test dataset, the main problem is that we cannot determine the T-shirt and Shirt well, as they look very similar, and for humans, they can not distinguish them well. 
\subsection{CIFAR10}
Furthermore, we apply our BackEISNN to the CIFAR10 dataset. Compared to MNIST and FashionMNSIT, CIFAR10 is a more challenging color image dataset with an image size of 32*32*3. It has 60,000 color images belong to 10 classes, and 50,000 for the train, 10,000 for the test. The dataset is normalized, random cropping, and horizontal flipping for data augmentation, commonly used for CIFAR10 preprocessing. We also used Dropout after each layer, which is also adopted in STBP and TSSL-BP. The network structure is 128C3-P2-256C3-P2-512C3-P2-1024. And the kernel size for the two modules is set with 3. The simulation length is charged with 20. 

As can be seen in Tab~\ref{cifar10}, the accuracy of our BackEISNN has reached 90.93 \%, which exceeds the SNNs trained with BP, decode Voting, and NeuNorm. For the TSSL-BP method, the warm-up mechanism is used that when very low firing activity levels are observed, the network will use the continuous sigmoid function of membrane potential to approximate the activation so that the errors can be propagated back when there is no spike for a deeper spiking neural network. And the warm-up mechanism is not adopted in our experiment. Also, our experiment uses a relatively light structure with three convolutional layers and two linear layers. In contrast, for the TSSL-BP and STBP, the deep network with five convolutional layers is used. 
\begin{table}[H]
	\centering
	\caption{Performances of BP based SNNs on CIFAR10}
	\begin{tabular}{cccc}
		\toprule
		Model & Method  &Performance\\
		\midrule
		STBP~\cite{wu2019direct} &BP and Decode Voting& 89.83 \\
		STBP~\cite{wu2019direct} &BP, Decode Voting and NeuNorm& 90.53 \\
		BackEISNN (Ours) &BP, SFBM and BEIM& \textbf{90.93}\\
		TSSL-BP~\cite{zhang2020temporal} &Inter-neuron BP and Intra-neuron BP  &91.41 \\
   
		\bottomrule
	\end{tabular}%
	\label{cifar10}%
\end{table}%
\begin{figure*}[htbp]
\centering
\includegraphics[width=18cm]{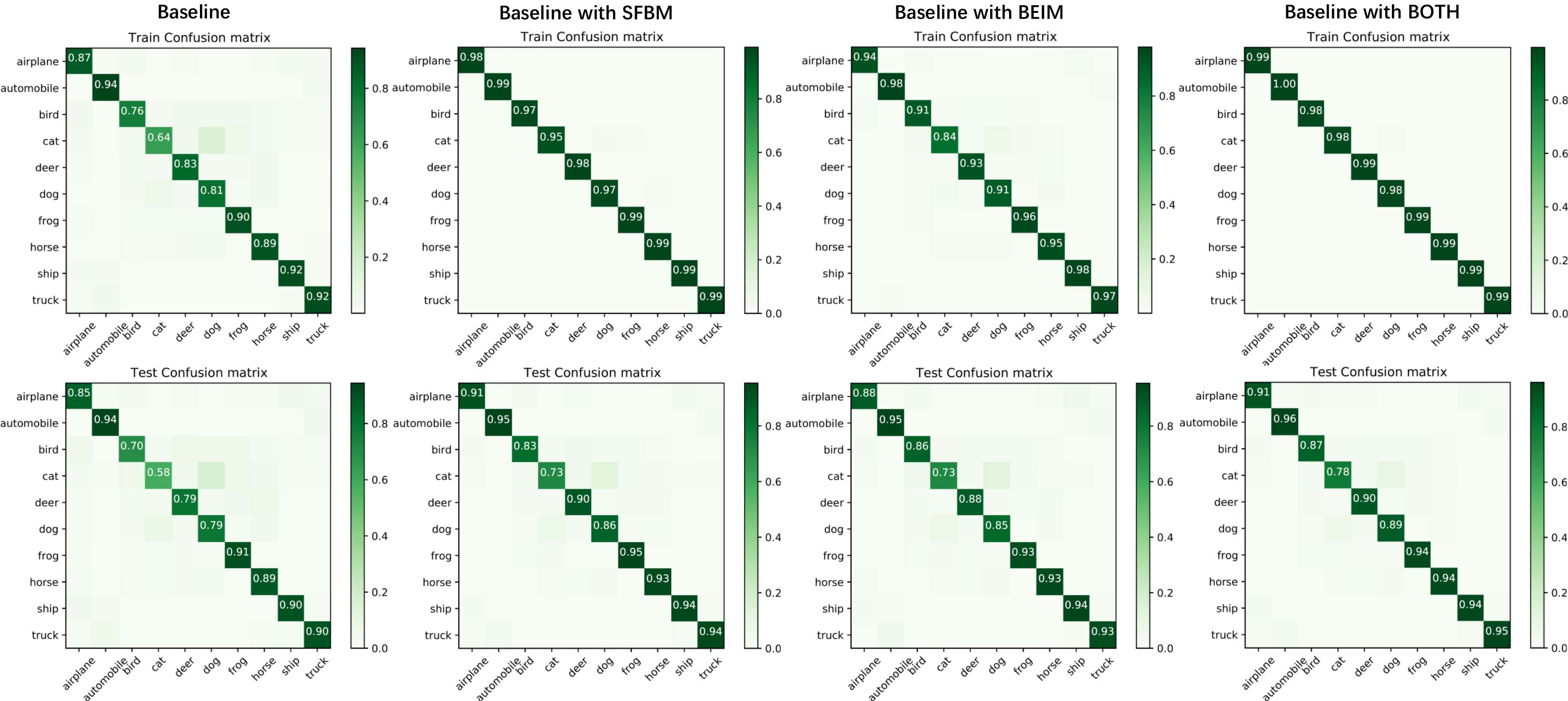}
\caption{Confusion matrix of the BP-based SNN with the SFBM, BEIM, and both of them on the CIFAR10 dataset. The top is the training dataset, and the bottom is the test dataset.}
\label{confusefinal}
\end{figure*}

\section{Discussion}
To further illustrate the contributions of different modules of our algorithm, we conducted ablation studies on the CIFAR10 experiment. The network structure is the same as mentioned above. As can be seen in Tab~\ref{tab:aba}, for the baseline without SFBM and BEIM, the accuracy is only 82.6\%. With the introduction of SFBM, the accuracy increases to 88.86\%, and with the introduction of BEIM, the accuracy rises to 89.32\%. With both of the mechanisms introduced, our BackEISNN has reached 90.93\% accuracy, which shows that the combination of the two mechanisms has better improved the network's performance.

\begin{table}[H]
	\centering
	\caption{Ablation Studies of BackEISNN on CIFAR10}
	\begin{tabular}{ccccc}
		\toprule
		Model & With SFBM  &With BEIM &Performance\\
		\midrule
		Baseline &- & -& 82.6 \\
		Baseline &\checkmark & -& 88.86 \\
		Baseline &- & \checkmark & 89.32 \\
		Baseline &\checkmark & \checkmark & 90.93 \\
		\bottomrule
	\end{tabular}%
	\label{tab:aba}%
\end{table}%

\begin{figure}[htbp]
	\centering
	\includegraphics[width=8cm]{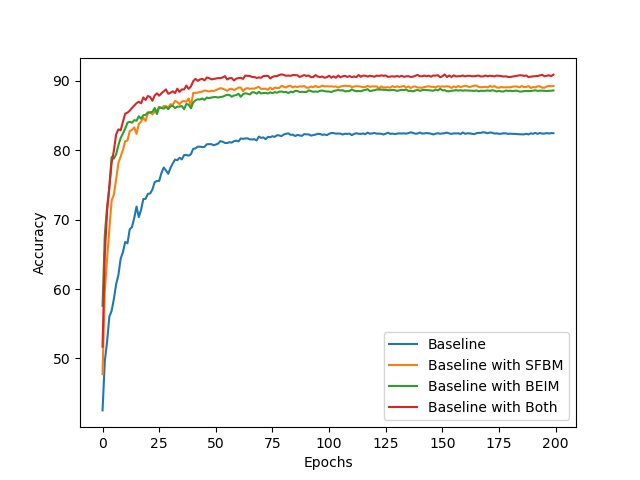}
	\caption{The test accuracy on CIFAR10 dataset with and without SFBM and BEIM.}
	\label{both}
\end{figure}

We also plot the baseline's test accuracy, baseline with SFBM, baseline with BEIM, baseline with Both. As can be seen in Fig~\ref{both}, with the introduction of two mechanisms, our network has not only higher accuracy but also a faster convergence rate. 

To better illustrate the two modules' superiority, we plot the confusion matrix of the four models. As shown in Fig~\ref{confusefinal}, for the original model without SFBM and BEIM module, for the similar object, such as airplane and bird, the cat and dog, the horse and deer, the networks would be confused to distinguish them clearly. However, when the two modules are introduced, the networks would distinguish them more clearly. For the other not confused categories, the BackEISNN also improves the accuracy. That's to say, with the introduction of the adaptive self-feedback module and the dynamic balance excitatory and inhibitory neurons, the network not only improves the accuracy of the easily classified categories but also improves the classification accuracy of the easily confused categories to a large extent. 
\section{Conclusion}
There is still a performance gap between the BP-based SNNs and DNNs due to the particular data transmission mode in the SNNs, indicating that only BP is not enough for the training. In this paper, inspired by the autapses, which connect the soma in a self-feedback manner, we propose to apply the time-delayed feedback on the neuron's membrane potential. The input current is gated with the spikes' information at the last moment to regulate the spike precision and the temporal information. Also, we adopt the balanced excitatory and inhibitory neuron mechanism to control the output form of the spiking neurons dynamically. The introduction of the two mechanisms into the spiking neural networks trained with BP accelerates the convergence of the network and improves the accuracy. For the MNIST, FashionMNIST, and N-MNIST, our BackEISNN has achieved state-of-the-art performance. For the CIFAR10 dataset, our BackEISNN also gets remarkable performance in a relatively light structure that competes against state-of-the-art SNNs.

\section*{Acknowledgement}
This work is supported by the National Key Research and Development Program (2020AAA0104305), the Strategic Priority Research Program of the Chinese Academy of Sciences (Grant No. XDB32070100), the Beijing Municipal Commission of Science and Technology (Grant No. Z181100001518006), and the Beijing Academy of Artificial Intelligence (BAAI). 

\ifCLASSOPTIONcaptionsoff
  \newpage
\fi



%
\bibliographystyle{IEEEtran}
\bibliography{IEEEabrv,mybibfile}



%

\begin{IEEEbiography}[{\includegraphics[width=1in,height=1.25in,clip,keepaspectratio]{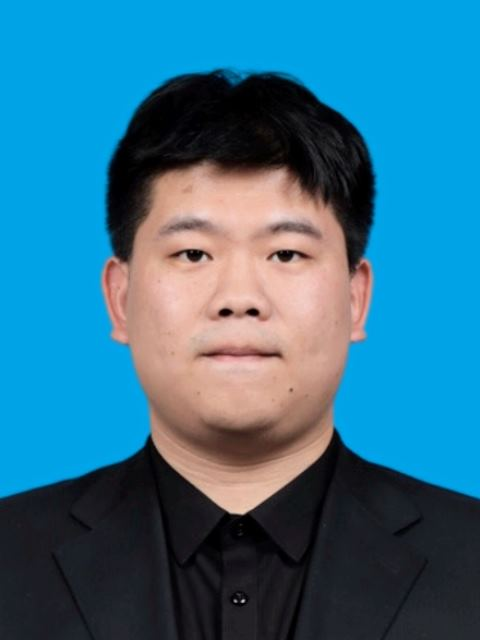}}]{Dongcheng Zhao}
received the B.Eng. degree in Information and Computational Science from XiDian University, Xi'an, Shaanxi, China, in 2016. He is currently pursuing the PH.D. degree with the Institute of Automation, Chinese Academic of Sciences, Bejing, China.
His current research interests include learning algorithms in spiking neural networks, thalamus-cortex interaction, visual object tracking, etc. 
\end{IEEEbiography}
\begin{IEEEbiography}[{\includegraphics[width=1in,height=1.25in,clip,keepaspectratio]{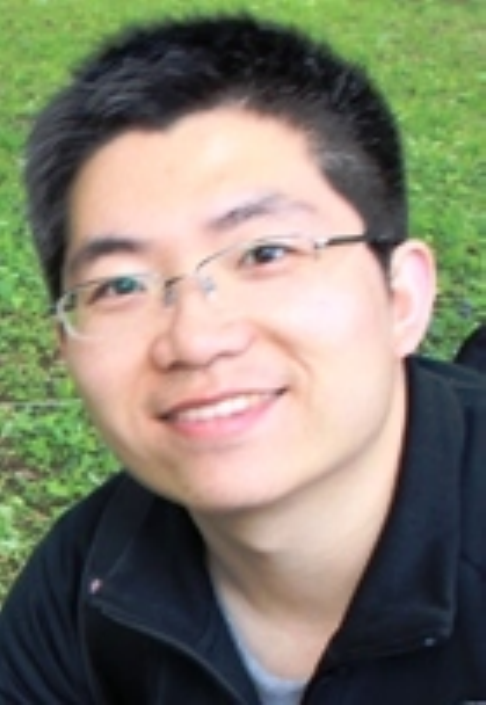}}]{Yi Zeng}
obtained his Bachelor degree in 2004 and Ph.D degree in 2010 from Beijing University of Technology, China. He is currently a Professor and Deputy Director at Research Center for Brain-inspired Intelligence, Institute of Automation, Chinese Academy of Sciences (CASIA), China. He is also with the National Laboratory of Pattern Recognition, CASIA, and University of Chinese Academy of Sciences, China. He is a Principal Investigator at Center for Excellence of Brain Science and Intelligence Technology, Chinese Academy of Sciences, China. His research interests include cognitive brain computational modeling, brain-inspired neural networks, brain-inspired robotics, etc.
\end{IEEEbiography}
\begin{IEEEbiography}[{\includegraphics[width=1in,height=1.25in,clip,keepaspectratio]{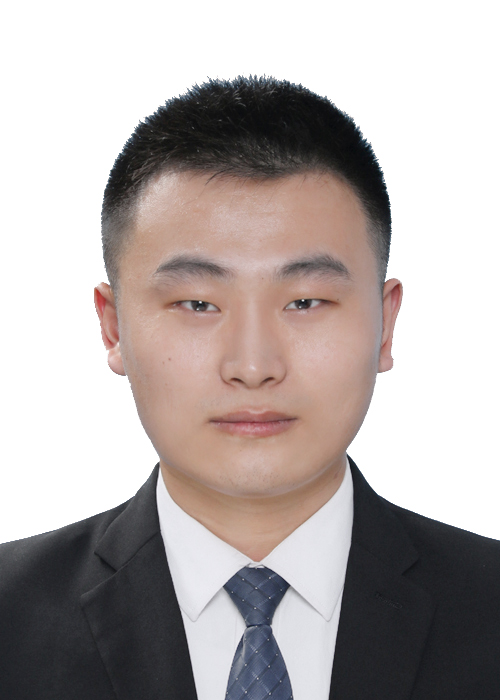}}]{Yang Li}
received the B.Eng. degree in automation from Harbin Engineering University, Harbin, China, in 2019. He is currently pursuing the master's degree with the Institute of Automation, Chinese Academic of Sciences, Bejing, China.
His current research interests include learning algorithms in spiking neural networks and cognitive computations.
\end{IEEEbiography}




\end{document}